\documentclass{article} 
\usepackage{iclr2023_conference,times}

\usepackage{amsmath,amsfonts,bm}

\def\eqref#1{equation~\ref{#1}}

\def\1{\bm{1}}

\DeclareMathAlphabet{\mathsfit}{\encodingdefault}{\sfdefault}{m}{sl}
\SetMathAlphabet{\mathsfit}{bold}{\encodingdefault}{\sfdefault}{bx}{n}

\usepackage{hyperref}
\usepackage{url}

\usepackage{graphicx}
\usepackage{multirow}
\usepackage{amsmath}
\RequirePackage{algorithm}
\RequirePackage{algorithmic}

\title{SaiT: Sparse Vision Transformers through Adaptive Token Pruning}

\author{Ling Li, David Thorsley, and Joseph Hassoun \\
Samsung Semiconductor Inc.\\
\texttt{\{ling.li, d.thorsley, j.hassoun\}@samsung.com} 
}

\iclrfinalcopy 
\begin{document}

\maketitle

\begin{abstract}
While vision transformers have achieved impressive results, effectively and efficiently accelerating these models can further boost performances. 
  In this work, we propose a dense/sparse training framework to obtain a unified model, enabling weight sharing across various token densities. Thus one model offers a range of accuracy and throughput tradeoffs for different applications.
  Besides, we introduce adaptive token pruning to optimize the patch token sparsity based on the input image. 
  In addition, we investigate knowledge distillation to enhance token selection capability in early transformer modules. 
  \textbf{S}parse \textbf{a}daptive \textbf{i}mage \textbf{T}ransformer (\textbf{SaiT}) offers 
  varying levels of model acceleration by merely changing the token sparsity on the fly. Specifically, SaiT reduces the computation complexity (FLOPs) by 39\% - 43\% and increases the throughput by 67\% - 91\% with less than 0.5\% accuracy loss for various vision transformer models. Meanwhile, the same model also provides the zero accuracy drop option by skipping the sparsification step. SaiT achieves better accuracy and computation tradeoffs than state-of-the-art transformer and convolutional models.  
\end{abstract}

\section{Introduction}
Even though Convolutional Neural Networks (CNNs) \citep{alex, resnet, efficientnet} have fueled rapid development in the computer vision field,  emerging studies on vision transformers show encouraging results, with some surpassing CNNs in a wide range of tasks such as classification \citep{ViT, deit, swin}, semantic segmentation \citep{maskformer, SETR} , and object detection \citep{detr, vitdet}. To improve model efficiency, especially on edge devices, model compression techniques such as pruning \citep{deepcompression}, quantization \citep{quantization0}, and knowledge distillation \citep{knowledgedistill} have been widely used in CNNs. However, model acceleration/compression in vision transformers is still less explored. Additionally, these typical compression techniques -- which usually lead to some accuracy loss -- are not ideal for accuracy-sensitive applications. 

\begin{figure}[hb]
  \centering
  \includegraphics[scale=0.42]{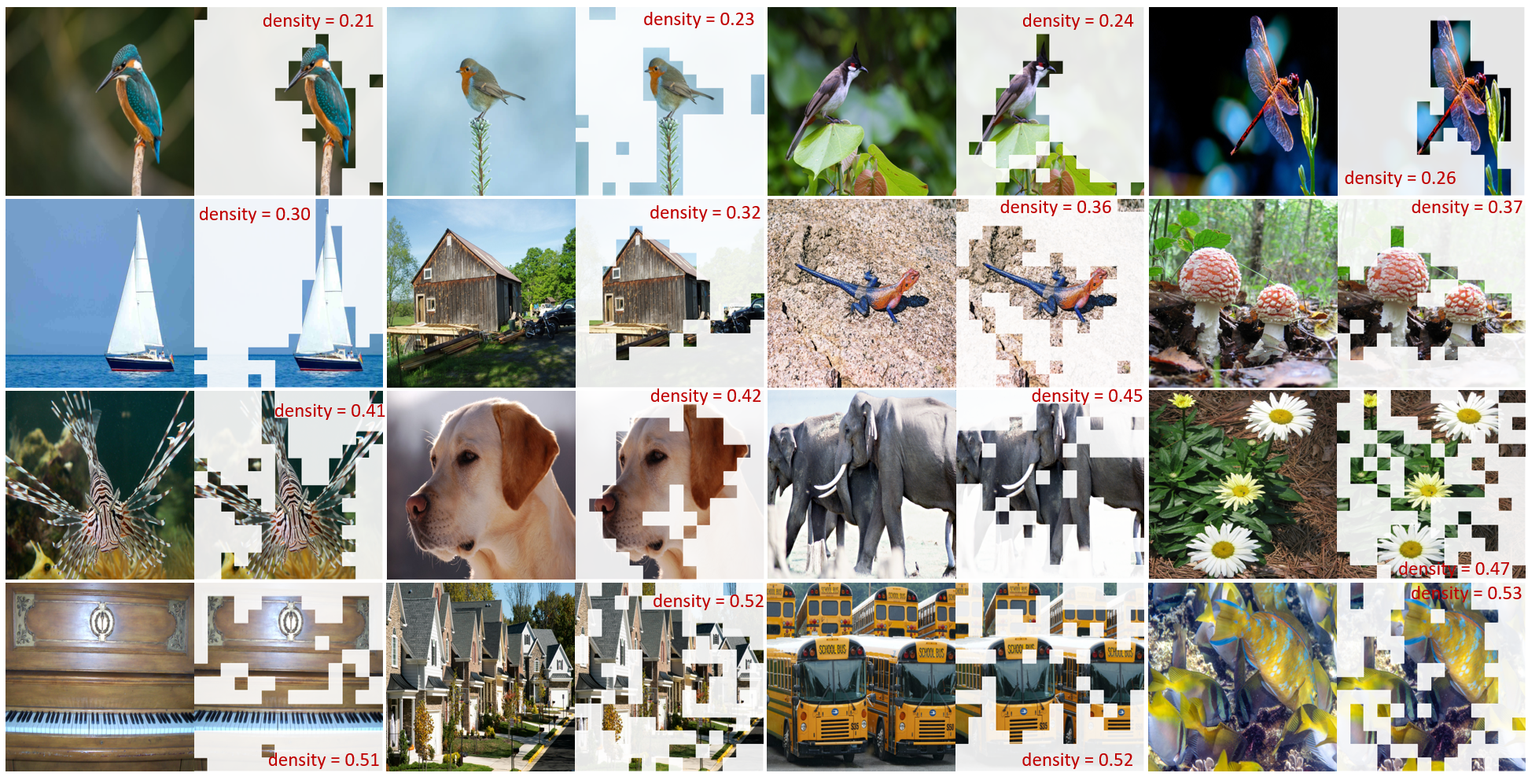}
  \caption{Visualization of the adaptive pruning based on the results from SaiT-S$^\dagger$. Original image and sparsification results are presented next to each other. Based on the difficulties of the inputs, the densities of essential patch tokens dynamically change from 21\% to 53\%.}
  \label{visualization}
\end{figure}

For efficient and hardware-friendly model acceleration, we leverage the intrinsic structure of vision transformers, where input images are transformed into patch tokens before further processing. Patch tokens allow the vision transformer to process the entire image; however, the computation for the all-to-all attention is expensive. Token pruning is an effective approach to reduce computation and save memory. The essential number of patch tokens varies depending on the input image, since background patches often contribute little for correct classification.  Some 'easy' inputs require fewer patches and 'difficult' inputs need more patches.  
Figure~\ref{visualization} shows that for 'easy' inputs, such as the bird image on the upper left, around 20\% of patches are sufficient for detection, whereas 'difficult' inputs, such as the fishes on the lower right, require 53\% token density. To save more computation based on the specifics of input images, we propose an adaptive token pruning strategy that dynamically adjusts the number of preserved tokens. This approach evaluates the importance (in probability) of each token based on the attention scores of the early transformer modules. Instead of selecting a fixed number of tokens, we accumulate a varying number of the most important tokens up to a probability mass threshold. As a result, this introduces no extra parameters and negligible computation, and thus its efficiency compares favorably with some prior works \citep{dynamic, dyvit}. 

In addition, we formulate a dense/sparse training framework to obtain a unified model which can flexibly adjust the tradeoff between accuracy and throughput on demand. Different computation savings are achieved by merely modifying the token density in the later transformer modules. Sharing the same weights throughout the transformer model, fully dense patch tokens preserve accuracy, but no model acceleration, while sparse tokens offer varying levels of acceleration in return for some accuracy drop. Therefore, different applications can share the same model and weights regardless of accuracy and throughput requirements. Consequently, this approach saves training cost and memory footprint by training and storing a single unified model instead of a series of different models.  

To demonstrate the effectiveness of this approach, we deploy our proposed training framework and sparsification schemes on top of DeiT \citep{deit} and LVViT \citep{lvvit}. The resulting unified model, \textbf{S}parse \textbf{a}daptive \textbf{i}mage \textbf{T}ransformer (\textbf{SaiT}), offers different levels of sparsification and achieves 39\% - 43\% floating point operation (FLOP) reduction and 74-91\% throughput gain with less than 0.5\% accuracy loss. In summary, we present three major contributions of this work: 
1) We formulate a dense/sparse training framework to obtain a unified model offering a range of accuracy and throughput tradeoffs;
2) We propose an adaptive pruning strategy to flexibly and dynamically adjust the token sparsity based on the input images; 3) We introduce knowledge distillation to improve the accuracy of early transformer modules in learning the token importance.

\section{Related Work}
\label{related work}

\textbf{Vision Transformers} ViT~\citep{ViT} is the first pure transformer based-model for image classification with results competitive to CNNs. However, the training of ViT requires a large private dataset JFT300M \citep{jft}. To address this issue, DeiT~\citep{deit} develops a training schedule with data augmentation, regularization, and knowledge distillation. Many subsequent variants ~\citep{cait, swin, twins, lvvit} of ViT and DeiT achieve promising performances, with some even surpassing CNN counterparts~\citep{resnet, efficientnet, regnet}. Moreover, self-supervised vision transformers, such as DINO \citep{dino} and MAE \citep{mae}, not only achieve impressive classification accuracy but also obtain useful feature representations for downstream tasks such as object detection and segmentation. 

\textbf{Efficient Vision Transformers} To improve model efficiency and save computation, \citet{centroid} and \citet{ perceiver} introduce new attention modules, while other works~\citep{localvit, botnet, levit, bossnet, cvt, coat, cmt} incorporate some convolutional layers into vision transformers. Following conventional model compression approaches, \citet{qvit} apply post-training quantization in vision transformers, \citet{svite} study the sparsity via sparse subnetwork training, and \citet{dyvit} implement a lightweight prediction module for hierarchical token pruning. \citet{dynamic} dynamically adjusts the number of patch tokens through cascading multiple transformers with confidence estimation modules. \citet{evit} exploits the classification token to select attentive tokens without introducing extra parameters and fuse inattentive tokens through multiple stages for efficient inference.  

Compared to prior works, our token pruning strategy is more efficient by leveraging the attention scores for one-stage adaptive pruning, with no extra parameters and negligible computation.
Besides, the unified model from our dense/sparse training framework offers a range of accuracy and throughput tradeoffs for different applications. 

\section{SaiT}
\label{sait}

The proposed dense/sparse training framework and adaptive token pruning apply to general vision transformer architectures. To illustrate the ideas, we use DeiT \citep{deit} and LVViT \citep{lvvit} as examples in this work. Both DeiT and LVViT use an embedding module to convert an input image into $N$ patch tokens. These $N$ patch tokens, along with the classification token $CLS$, go through a series of transformer modules/layers. The feature representation from the last transformer layer is used for the final classification. The key to the proposed approach is to enable early layers to effectively capture the importance of each token, in order to reduce computation in the later layers with sparse tokens. 

\subsection{Dense/sparse training framework}

\begin{figure}[ht]
  \centering
  \includegraphics[scale=0.64]{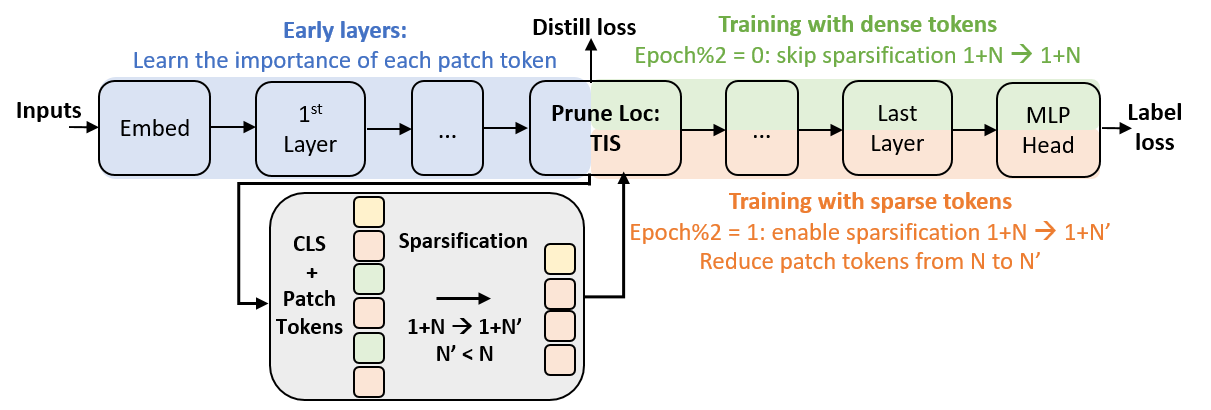}
  \caption{An overview of the dense/sparse training framework used in SaiT. Early layers (blue) learn the importance of each token. The attention scores at the prune-location are used to extract token importance score ($TIS$) and the token mask. Later layers are trained alternately with fully dense tokens (green) and sparse tokens (orange). Optionally, knowledge distillation from a teacher model enhances $TIS$ learning ability of early layers.}
  \label{framework}
\end{figure}

The overview of the dense/sparse training framework is in Figure~\ref{framework} and Algorithm~\ref{train_alg}. Given an architecture with $L$ total transformer layers, the early layers ($l^0$ to $l^{P-1}$) learn to identify the importance of each patch token. At the designated pruning location ($l^P$), token importance scores ($TIS$) are extracted based on the attention scores, which are used for token selection and knowledge distillation. Later layers ($l^{P+1}$ to $l^{L-1}$) are trained on alternate epochs with $N$ fully dense patch tokens (without pruning) and $N'$ sparse patch tokens (after pruning).  

\begin{algorithm}
   \caption{A General Training Framework used in SaiT}
   \label{train_alg}
\begin{algorithmic}
   \STATE {\bfseries Input:} pretrained teacher model $W_t$, token pruning location $l^P$, distillation ratio $\beta$
   \STATE Initialize student weights $W_s$.
   \FOR{epoch $i=1$ {\bfseries to} $T$}
   \STATE Obtain the classification token $\overline{CLS}$ from the last layer of the teacher model 
   \STATE Obtain student $TIS^{*P}$ and token mask $M$ at layer $l^P$,
   \STATE Calculate token distillation loss $L_{distill} = KL(TIS^{*P},\overline{CLS}) $ 
   \IF{$i \mod  2 = 0$}
        \STATE \textbf{Dense} -- skip token sparsification and use all the patch tokens after layer $l^P$ 
   \ELSE
        \STATE \textbf{Sparse} -- prune tokens based on the token mask $M$ and use only sparse tokens after layer $l^P$
   \ENDIF
   \STATE Compute the label loss $L_{label}$ and total loss $L = L_{label} + \beta L_{distill}$
   \STATE Update student network
   \ENDFOR
\end{algorithmic}
\end{algorithm}

\textbf{Dense/sparse alternate training} This training schedule enables weight sharing between fully dense (unpruned) and sparse (pruned) patch tokens at layers $l^{P+1}$ to $l^{L-1}$, since the weights of transformer blocks are independent of the number of patch tokens. Moreover, it improves processing accuracy of later layer on sparse tokens as shown in the ablation study (Section~\ref{ablation_sec}). 
Besides, training with this framework preserves the model accuracy when skipping the sparsification. This is different from prior pruning models \citep{dyvit, evit}, which are unable to recover the original accuracy with dense tokens. 

\subsection{Token sparsification strategies}

We leverage the attention scores to extract the Token Importance Score ($TIS$) for token sparsification during training and inference. For the token $n$, at the pruning layer $l^P$, $TIS$ is defined as:
\begin{equation} 
\label{TIS}
TIS_{n}^P = 
    \frac 
    {W_{n}^P} 
    {\sum_{i=0}^N W_i^P}, 
    W_{n}^P = 
    \sum_{h=0}^H 
    \sum_{m=0}^N attn_{h, m, n}^P
\end{equation}
where $attn_{h, m, n}^P$ is the attention score at head $h$,  row $m$, column $n$, layer $l^P$, and $attn = Softmax\left( \frac {QK^T} {\sqrt{d}} \right)$.

We skip performing softmax and averaging over all the heads to simplify computation. 
Intuitively, a patch token $n$ is more important if it contributes heavily across all rows when computing $attn \times V$. Therefore summation across all rows of $attn$ at column $n$ approximately reflects its the importance.  

We derive two strategies based on $TIS$: \textbf{value-based} and \textbf{mass-based} sparsification schemes.

\textbf{A. Value-based Token Selector ($TS_V$)} In the value-based sparsification scheme, we select a fixed number of tokens ($K$) with the greatest $TIS$ values:
\begin{equation} 
\label{value-based}
    TS_V = top_K (TIS_{n\in \{1, \dots, N\}}^P)
\end{equation}

For a given target token density $\rho$, we have $K$ = round$(\rho * (N+1))$. 

\textbf{B. Mass-based Token Selector ($TS_M$)} Based on the distribution of $TIS$, we select the minimum number of highest weighted tokens whose probability sum up to or above the mass threshold $M_{th}$: 

\begin{equation} 
\label{mass-based}
    TS_M = top_S (TIS_{n\in \{1, \dots, N\}}^P),\: s.t.\: \min_{S} \sum_{i \in TS_M} TIS_i^P >= M_{th}
\end{equation}

Intuitively (and as shown in Figure~\ref{visualization}), patches containing target objects have higher $TIS$ and the background patches have lower $TIS$. When small target objects occupy fewer patches, the corresponding distribution of $TIS$ is typically more concentrated, whereas large objects spread their $TIS$ over larger area. As a result, for a given mass threshold $M_{th}$, $TS_M$ is able to adjust the number of selected tokens according to the image input. 

\textbf{Batch Training (with $TS_M$)}
To accommodate the varying number of tokens from mass-based pruning, we convert $TS_M$ to a binary token mask $M$: 

\begin{equation} 
\label{mask}
    M_{j} = 
    \begin{cases}
        1   &j \in TS_M \\
        0   &j \notin TS_M
    \end{cases},
    \forall j \in \{1, \dots, N\}
\end{equation}

Accordingly, we modify the attention module to perform the all-to-all attention only on the remaining tokens after sparsification, by setting attention products related to pruned tokens to negative infinity: 
\begin{equation} 
\label{attn}
    QK^T = 
    \begin{cases}
        QK^T_{h, m, k}= -\infty   & \text{if } M_m = 0 \text{ for all } k \\
        QK^T_{h, j, n}= -\infty   & \text{if } M_n = 0 \text{ for all } j\\
        QK^T_{h, m, n}          & \text{if } M_m = 1 \text{ and } M_n = 1
    \end{cases}
\end{equation}
Where $QK^T$ is the product of Query ($Q$) and Key ($K$) to compute $attn = Softmax(QK^T / \sqrt{d})$. The elements associated with the selected tokens remains the same while those of the pruned tokens are set to negative infinity (in practice as -65,000). This sets all the attention probabilities (columns and rows) corresponding to pruned tokens to zero. Subsequently, the pruned tokens are all zeros in the feature maps ($attn \times V$).

\subsection{Knowledge distillation} 
Optionally, we introduce knowledge distillation at $l^P$ from a teacher model to improve the ability of early layers to learn $TIS$. For DeiT, we choose a self-supervised vision transformer -- DINO \citep{dino} as the teacher, since DINO and DeiT share the same model architecture. Furthermore, DINO contains semantic segmentation information of an image by allocating the top attention in the classification token on the target object, and thus it serves as an ideal teacher to identify $TIS$. 

For distillation loss during training, $TIS^*$ of token $n$ at pruning layer $l^P$ is slightly modified as following, in order to match the distribution in the teacher $\overline{CLS}$: 

\begin{equation} 
\label{tis_c}
TIS_n^{*P}=\frac 1 H \sum_{h=0}^H 
    \left( 
    Softmax
    \left(
    \sum_{m=0}^N attn_{h, m, n}^P
    \right)
    \right)
\end{equation}

The distillation loss is computed through KL divergence: 

\begin{equation} 
\label{distill loss}
    L_{distill} = KL
    \left( 
    TIS^{*P} \| \overline{CLS}
    \right)
\end{equation}
Where $\overline{CLS}$ is the classification token from the last layer of DINO backbone, averaged across all attention heads.


Combined with the label loss, the final training objective is as follows: 
\begin{equation} 
\label{loss}
    L_{tot} = L_{label} + \beta L_{distill}
\end{equation}
Where $L_{label}$ = CrossEntropy($y, \bar y$), as the cross-entropy loss between the model predictions $y$ and the ground truth labels $\bar y$. $\beta$ is the distillation ratio. 

Note LVViT \citep{lvvit} uses TokenLabeling, which offers additional supervision and is beneficial for the token selection. Subsequently, for LVViT, we have options to skip knowledge distillation or to use a pre-trained LVViT as the teacher.  

\section{Experimental Results}
\label{experiments}

\begin{figure}[ht]
\begin{minipage}{.48\textwidth}
  \includegraphics[width=1.\linewidth]{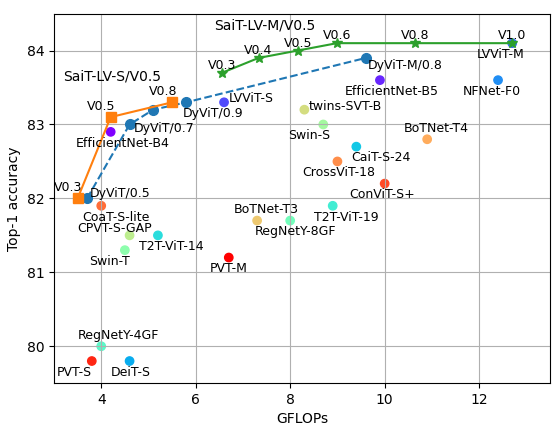}
  \caption{Comparison of SaiT-LV with SOTA models. 
  Green and orange lines represents unified models of SaiT-LV-M and SaiT-LV-S respectively. SaiT-LV is trained end-to-end once and inferred at various token densities (0.3 to 0.8). SaiT-LV outperforms the existing SOTA models with better accuracy-computation(FLOPs) tradeoffs.}
  \label{sota}
\end{minipage} \hfill%
\begin{minipage}{0.48\textwidth}
  \centering
  \includegraphics[width=1.\linewidth]{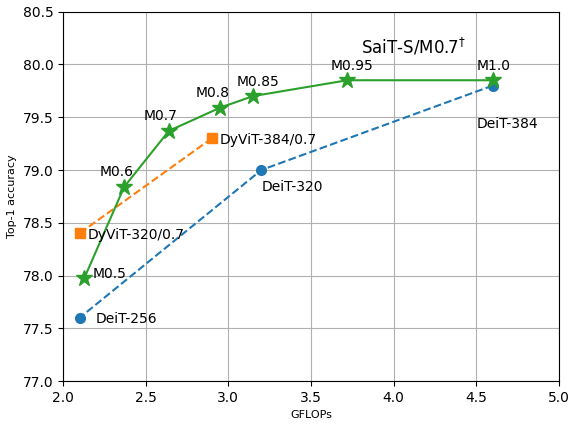}
  \caption{Accuracy-computation (FLOPs) tradeoffs of the SaiT-S/M0.7$^\dagger$ model. Green line represents SaiT-S$^\dagger$ inferred at various mass thresholds on the fly. Orange and blue dash lines connect multiple DyViT and DeiT models with various hidden dimensions. A single SaiT-S$^\dagger$ model offers a range of accuracy-computation results outperforming a series of DeiT models.}
  \label{scalability}
\end{minipage}
\end{figure} 

\begin{table}[ht]
    \caption{Comparison of top-1 accuracy between SaiT and some SOTA models.}
    \label{sota accuracy}
    \centering
    \begin{tabular}{llll}
    \\
        \textbf{Models}   & \textbf{Params }(M)    & \textbf{GFLOPs}    & \textbf{Top-1 Acc.} \\  
        \hline
        PVT-S \citep{pvt}        &24.5    &3.8    &79.8\\
        DeiT-S \citep{deit} & 22.1  &4.6    &79.8 \\
        CrossViT-S \citep{crossvit}     &26.7   &5.6    &81.0\\
        Swin-T \citep{swin}             &29        &4.5  &81.3 \\ 
        T2T-ViT-14 \citep{t2t}        &21.5      &5.2  &81.5 \\
        Twins-SVT-S \citep{twins}     &24       &2.8    &81.7\\
        CaiT-XS \citep{cait}         &26.6    &5.4  &81.8 \\
        CoaT-Lite-S \citep{coat}    &20     &4.0    &81.9 \\
        DeepViT-S \citep{deepvit}     &27    &6.2  &82.3 \\
        EfficientNet-B4 \citep{efficientnet}    &19.3   &4.2    &82.9 \\
        \hline
        
        LVViT-S \citep{lvvit}   &26.2   &6.6    &83.3 \\
        DyViT-LV-S/0.7 \citep{dyvit}    &26.9   &4.6    &83.0 \\
        \textbf{SaiT-LV-S/V0.5}           &26.2   &4.3    &83.1 \\
        
        \hline
        T2T-ViT-24 \citep{t2t}     &64.1      &14.1  &82.3 \\
        CaiT-S \citep{cait}     &46.9      &9.4  &82.7 \\
        Swin-S \citep{swin}     &50      &8.7  &83.0 \\ 
        Twins-SVT-B \citep{twins}      &56       &8.3    &83.2 \\
        EfficientNet-B5 \citep{efficientnet}    &30.0   &9.9    &83.6 \\
        NFNet-F0 \citep{nfnet}  &72     &12.4    &83.6 \\
        \hline 
        LVViT-M \citep{lvvit}   &55.8   &12.7    &84.1 \\
        DyViT-LV-M/0.7 \citep{dyvit}    &57.1   &8.5    &83.8 \\
        \textbf{SaiT-LV-M/V0.5 }          &55.8   &8.2    &84.0 \\
        
    \end{tabular}
\end{table}

In this section, we compare the performances of SaiT with various state-of-the-art (SOTA) vision transformer models, as well as representative CNN models. We explore the accuracy and throughput tradeoffs in the unified model. In addition, we analyze the impact of dense/sparse alternate training and knowledge distillation through an ablation study. The performance impacts of various factors are also discussed, such as sparsification strategies, mass thresholds, and distillation teachers.

\subsection{Setup}

SaiT-S and SaiT-LV share the same architectural configurations and training schedules as DeiT-S \citep{deit} and LVViT \citep{lvvit}, respectively. The implementation is based on the Timm library \citep{timm}. ILSVRC-2012 ImageNet \citep{imagenet} is used for training and evaluation with the input resolution as 224$\times$224. We train SaiT-S and SaiT-LV from scratch with the proposed training framework for 300 epochs and 410 epochs, respectively. The default batch size is 1024 on 4 GPUs. The default distillation ratio ($\beta$) is 4. For ease of notation, we use SaiT/M0.7 and SaiT/V0.5 to represent the models trained with a mass-based threshold of 0.7 and with a value-based token density of 0.5, respectively. The default pruning location is Layer 3 for SaiT-S and SaiT-LV-S, and Layer 5 for SaiT-LV-M. During training, the default mass threshold and token density are 0.7 and 0.5. For inference, the mass threshold and token density change between 0 to 1 on the fly, subject to the accuracy and throughput requirements. The models with and without distillation are denoted as SaiT$^\dagger$ and SaiT respectively.



\subsection{Comparison with SOTA}

As shown in Figure~\ref{sota}, SaiT-LV offers the best accuracy for given computation budgets (FLOPs) than other SOTA models. One unified model, SaiT-LV-M/V0.5, offers a range of accuracy 83.7\% - 84.1\% with 6.6 - 12.6 GFLOPs when inferred at token densities of 0.3 - 1.0. The computation efficiency of SaiT-LV outperforms not only transformer-based architectures but also CNN-based models such as EfficientNet\citep{efficientnet}, RegNet\citep{regnet}, and NFNet\citep{nfnet}.  
Specifically, as shown in Table~\ref{sota accuracy}, SaiT-LV-S/V0.5 and SaiT-LV-M/V0.5 achieve 1.8\% and 1.0\% higher accuracy than Swin-T and Swin-S, with 0.2G and 0.5G fewer FLOPs respectively.     
Moreover, SaiT-LV-S/V0.5 and SaiT-LV-M/V0.5 surpass EfficientNet-B4 and EfficientNet-B5 by 0.2\% and 0.4\% in top-1 accuracy with the same and 1.8G fewer FLOPs, respectively. 

\subsection{Accuracy and throughput tradeoff}


\begin{table}[ht]
  \caption{Comparison of throughput improvements between SaiT, DyViT \citep{dyvit}, and EViT \citep{evit} over baseline models. Throughput is measured on one 48GB RTX Quadro-8000 GPU with a batch size of 64, following the same procedure in \citep{swin}. $^1$SaiT-S/M0.7$^\dagger$ uses averaged token density and GFLOPs.}
  \label{throughput}
  \centering
  \begin{small}
  \begin{tabular}{llllll}
    \\ 
    \textbf{Models}      & \textbf{Top-1 Acc.}    & \textbf{Params} (M)    & \textbf{Token Density}    & \textbf{GFLOPs}    &\multicolumn{1}{p{1cm}}{\textbf{Throughput} (im/s)} \\
    \hline
    DeiT-S        & 79.8    & 22.1     & 1.0    & 4.6    &740.7 \\
    DyViT-S/0.7   & 79.3 (-0.5)    & 22.8 (+0.7)   & [0.7/0.49/0.34]        & 2.9 (-37\%)    & 1155.6 (+56\%) \\
    EViT-S    & 79.5 (-0.3)    & 22.1 (-)      & [0.7/0.49/0.34]  & 3.0 (-35\%)    & 1146.7 (+55\%) \\
    SaiT-S/M0.7$^\dagger$  & 79.4 (-0.4)   & 22.1 (-)     & 0.42$^1$   & 2.6 (-43\%)$^1$     & 1413.6 (+91\%) \\
    \hline
    LVViT-S     & 83.3   & 26.2     & 1.0     & 6.6  &614.0\\
    DyViT-LV-S/0.7      & 83.0 (-0.3)    & 26.9 (+0.7)   & [0.7/0.49/0.34]     & 4.6 (-31\%)   & 820.5 (+34\%) \\
    SaiT-LV-S/V0.5      & 82.9 (-0.4)     & 26.2 (-)     & 0.42    & 4.0 (-39\%)    & 1025.1 (+67\%) \\
    \hline
    LVViT-M     & 84.1    & 55.8     & 1.0    & 12.7    &340.5 \\
    DyViT-LV-M/0.7  & 83.8 (-0.3)    & 57.1 (+1.3)   & [0.7/0.49/0.34]     & 8.5 (-33\%)   & 493.1 (+45\%) \\
    SaiT-LV-M/V0.5    & 83.9 (-0.2)   & 55.8 (-)     & 0.4   &7.4 (-42\%)      &581.5 (+71\%)\\
  \end{tabular}
  \end{small}
\end{table}

SaiT adopts one-stage pruning with no extra token selection modules, as opposed to three-stage sparsification in DyViT \citep{dyvit} (with explicit token selection modules) and EViT \citep{evit}. Additional pruning stages and token selection modules increase the computation, processing time, and memory footprint. As a result, SaiT achieves better accuracy and throughput tradeoff with fewer parameters. Specifically, compared to the baseline DeiT-S and LVViT, SaiT boosts the throughput by 67-91\% with around 40\% token density and less than 0.5\% accuracy loss, as shown in Table~\ref{throughput}. This throughput gain is 26-35\% higher than DyViT and EViT with similar accuracy.

Furthermore, SaiT-S covers a range of accuracy and throughput when inferred at different mass thresholds. As shown in Figure~\ref{scalability}, SaiT-S/M0.7$^\dagger$ outperforms DeiT-256, DeiT-320, DeiT-384, as well as the pruned models DyViT-384/0.7 at various computation complexities (FLOPs). Compared to training and storing a series of different models, SaiT saves training cost and memory footprint with one unified model. 

\subsection{Weight sharing across various token densities}
As shown in Table~\ref{evit-s vs sait-s}, one SaiT-S/M0.7$^\dagger$ model achieves 0.1\% to 0.5\% higher accuracy and 7\% to 19\% higher throughput than a series of EViTs trained separately at various keep rates (token densities). 
Moreover, training a unified model which is optimized simultaneously for various token sparsities, reduces training cost and memory footprint. 
Therefore, even with the additional costs from the knowledge distillation, overall SaiT saves training costs by training a single model once as opposed to training a series of models individually. 

\begin{table}[ht]
  \caption{Performance comparison between EViT-S \citep{evit} and SaiT-S. 
  Density$^1$ is averaged across the validate set and GFLOPs$^1$ is obtained based on the averaged density. Throughput is measured on one 48GB RTX Quadro-8000 GPU with a batch size of 64, following the same procedure in \citep{swin}. Note EViT-S* are five separated models trained individually with keep rates from 0.5 to 0.9.}
  \label{evit-s vs sait-s}
  \centering
  \resizebox{\textwidth}{!}{
  \begin{tabular}{cccc|cccc}
    \\
    \multicolumn{1}{c}{\bf Keep rate} 
    &\multicolumn{1}{c}{\bf GFLOPs}
    &\multicolumn{1}{c}{\bf Top-1 Acc.} 
    &\multicolumn{1}{c|}{\bf Throughput} 
    
    &\multicolumn{1}{c}{\bf $M_{th}$/Density$^1$} 
    &\multicolumn{1}{c}{\bf GFLOPs$^1$}
    &\multicolumn{1}{c}{\bf Top-1 Acc.} 
    &\multicolumn{1}{c}{\bf Throughput} 
    
    \\
    \hline
    DeiT &4.6 &79.8 &740.7 &DeiT &4.6 &79.8 &740.7 \\
    \hline
    \multicolumn{4}{c|}{\bf EViT-S*} & \multicolumn{4}{c}{\bf SaiT-S/M0.7$^\dagger$} \\
    \hline
    
    0.9     &4.0 (-13\%)    & 79.8 (-0.0) &858 (+16\%)   
    &0.95/0.74  &3.7 (-19\%)& \textbf{79.8 (-0.0)}  &\textbf{999 (+35\%)}\\
    
    0.8   &3.5 (-24\%)      & 79.8 (-0.0) &993 (+34\%)
    &0.92/0.68 &3.5 (-24\%) & \textbf{79.8 (-0.0)}   &\textbf{1060 (+43\%)} \\
    
    0.7   &3.0 (-35\%)      & 79.5 (-0.3) &1146 (+55\%)  
    &0.8/0.52 &3.0 (-35\%)  &\textbf{79.6 (-0.2)} &\textbf{1264 (+71\%)}  \\
     
    0.6   &2.6 (-43\%)      & 78.9 (-0.9) &1315 (+78\%) 
    & 0.7/0.42 &2.6 (-43\%) & \textbf{79.4 (-0.4)}   & \textbf{1414 (+91\%)}\\
    
    0.5   &2.3 (-50\%)      & 78.5 (-1.3) &1486 (+100\%) 
    &0.6/0.34  &2.4 (-48\%) & \textbf{78.8 (-1.0)}   & \textbf{1543 (107\%)}\\
    
  \end{tabular}
  }
\end{table}

\subsection{Ablation study}
\label{ablation_sec}

\textbf{Dense/sparse alternate training and DINO distillation} 
As shown in Table \ref{ablation}, for SaiT-S/V0.5, the naive approach to  train later layers solely with 50\% sparse tokens results in 1.0\% accuracy loss.
Training early layers with a DINO teacher alone improves the model accuracy only with dense tokens, but not for sparse tokens. 
Subsequently, the most effective approach is to combine dense/sparse alternate training and DINO distillation, improving the accuracy of both dense and sparse tokens simultaneously. The resulting model preserves the baseline accuracy when skipping sparsification and has a 0.4\% accuracy loss when processing only 50\% of the patch tokens.  

\begin{table}[ht]
  \caption{Impact of dense/sparse alternate training and DINO distillation for SaiT-S/V0.5$^\dagger$}
  \label{ablation}
  \centering
  \begin{small}
  \begin{tabular}{lllll}
    \\
    
    \textbf{Pruning} & \textbf{Alternate training} & \textbf{DINO Distillation} & \textbf{Acc. w/100\% tokens} & \textbf{Acc. w/ 50\% tokens} \\
    \hline
    $\times$  & $\times$    & $\times$    & 79.8  & 77.9 (-1.9) \\
    \checkmark  & $\times$    & $\times$    & 78.8 (-1.0)   & 78.8 (-1.0) \\
    $\times$    & $\times$  & \checkmark    & 80.1 (+0.3)   & 77.8 (-2.0) \\
    \checkmark  & $\times$    & \checkmark    & 79.2 (-0.6)   & 79.2 (-0.6) \\
    \checkmark  & \checkmark    & $\times$    & 79.7 (-0.1)   & 79.2 (-0.6) \\
    \checkmark  & \checkmark    & \checkmark    & \textbf{79.8 (--)}   &\textbf{79.4} (-0.4)\\
  \end{tabular}
  \end{small}
\end{table}

\begin{figure}[ht]
  \centering
  \includegraphics[scale=0.48]{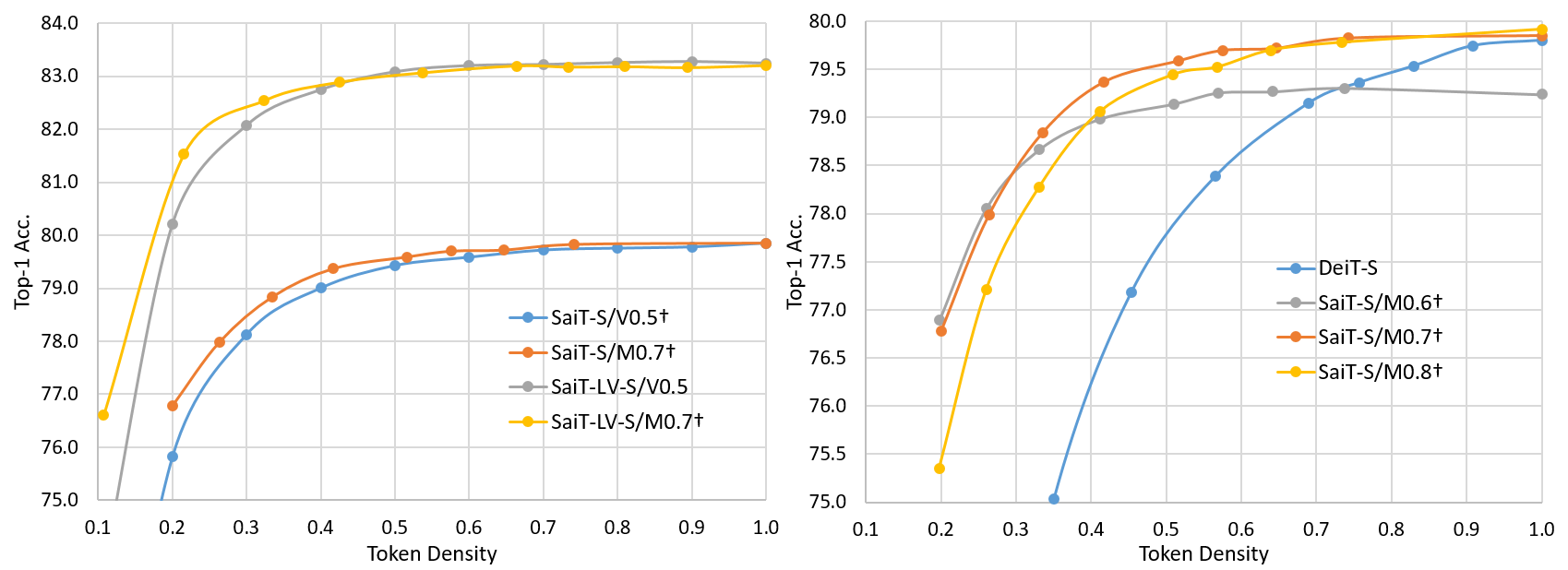}
  \caption{a) Top-1 accuracy for SaiT-S and SaiT-LV-S with mass-based and value-based sparsification strategies inferred at various thresholds and token densities. b) Top-1 accuracy of SaiT-S trained with different mass thresholds (0.6, 0.7, and 0.8), inferred at various thresholds (averaged token densities).}
  \label{sparsification}
\end{figure}

\textbf{Different sparsification strategies}
We study the impact of sparsification strategies in Figure~\ref{sparsification} a).
For both SaiT-S and SaiT-LV-S, mass-based strategy outperforms value-based strategy when token densities is below 0.5. This is attributed to the mass-based adaptive pruning, which adjusts the number of tokens according to the input images. Token densities for mass-based strategy are averaged across all the validation-set images. Note SaiT-LV-S/M0.7$^\dagger$ starts pruning at Layer 4 with more discussion in supplementary materials.

\textbf{Different mass thresholds}
We also evaluate the impact of different mass thresholds for training SaiT-S$^\dagger$ (Figure~\ref{sparsification} b). Compared to the model with mass threshold of 0.7, training with a smaller threshold of 0.6 leads to inferior inference accuracy at high token densities (0.3 - 1.0), whereas a larger threshold of 0.8 results in more accuracy loss at the low token density regime ($<$0.5) .  
We hypothesize that a smaller threshold reduce the number of available patch tokens to train later layers, and a large threshold is less effective to train the model with low token density. As a result, the mass threshold is set to be 0.7, for optimal inference accuracy of a wide range of token densities (0.2 - 0.9).  


\textbf{Different distillation teachers} To evaluate the impact of distillation teachers, we compare the SaiT-S$^\dagger$ performances with knowledge distillation from pre-trained DeiT-S and DINO.  
As shown in Table~\ref{distillation}, DINO as the teacher outperforms pre-trained DeiT-S for both dense and sparse tokens by 1.6\% and 2.3\% in top-1 accuracy respectively. This is because DeiT-S tends to allocate some large $TIS$ values to the background patches whereas DINO focuses more effectively on the target object \citep{dino}.

\begin{table}[h]
  \caption{Impact of different distillation teachers}
  \label{distillation}
  \centering
  \begin{tabular}{ccc}
    \\
    \textbf{Teacher model} & \textbf{Acc. w/100\% tokens}  & \textbf{Acc. w/50\% tokens}  \\
    \hline
    DeiT-S  & 78.2    & 77.1   \\
    DINO    & 79.8   & 79.4  \\
  \end{tabular}
\end{table}




\section{Conclusion}

In this work, we introduce a unified model with a range of accuracy and throughput tradeoffs for different applications. This is achieved through the dense/sparse training framework, knowledge distillation, and parameter-free token sparsification. The adaptive pruning scheme optimizes the token sparsity based on the difficulties of the inputs. SaiT preserves original accuracy when skipping sparsification, and achieves different levels of acceleration by changing token density on the fly. For various vision transformer architectures, the proposed approach effectively reduces 39-43\% FLOPs while increasing the throughput by 67-91\% with less than 0.5\% accuracy drop. Future works can extend the proposed methodologies to transformer/CNN hybrid models and downstream tasks.     



\bibliography{iclr2023_conference}
\bibliographystyle{iclr2023_conference}

\appendix
\section{Appendix}

\subsection{Token Selection: TIS vs. CLS}

We compared the performances of SaiT-S/V0.5 using TIS and CLS for token selection (Table~\ref{token selector}) since the CLS token is an alternative to TIS for token selection. However, when training with the CLS selector, the final model results in 0.3\% lower accuracy than the model based on TIS when inferred at dense (100\%) and sparse (50\%) tokens. 

\begin{table}[ht]
  \caption{Impact of different token selection methods: CLS vs. TIS on SaiT-S/V0.5}
  \label{token selector}
  \centering
  \begin{tabular}{ccc}
    \\
    \bf Token selection & \bf Acc. w/100\% tokens  & \bf Acc. w/50\% tokens  \\
    \hline
    TIS    & 79.7   & 79.2   \\
    CLS    & 79.4   & 78.9  \\
    
  \end{tabular}
\end{table}

\subsection{Distillation ratio}

To find the optimal distillation ratio for training, we test the values $\beta \in \{2, 4, 6\}$ on SaiT-S/V0.5. The resulting top-1 accuracies with dense (100\%) tokens and sparse (50\%) tokens are listed in Table~\ref{distillation_ratio}. Different distillation ratios result in no difference for 50\% sparse tokens. Using a distillation ratio of 4 leads to 0.1\% higher accuracy for 100\% dense tokens. Based on SaiT-S/V0.5, top-1 accuracy of both dense and sparse tokens is relatively insensitive to the distillation ratio over this range. As a result, we use $\beta=4$ as the default value in this work.  

\begin{table}[h]
  \caption{Top-1 accuracy of SaiT-S/V0.5$\dagger$ trained with various distillation ratios}
  \label{distillation_ratio}
  \centering
  \begin{tabular}{ccc}
    \\
    \bf distillation ratio ($\beta$) & \bf Acc. w/100\% tokens  & \bf Acc. w/50\% tokens  \\
    \hline
    2  & 79.8    & 79.4   \\
    4   & 79.9   & 79.4  \\
    6   & 79.8   & 79.4  \\
    
  \end{tabular}
\end{table}

\subsection{Comparison between EViT and SaiT}

Similarly, SaiT-LV-S/V0.5 achieves 0.1\% and 0.2\% higher accuracy than EViT as shown in Table~\ref{evit-lv-s vs. sait-lv-s}. The throughput improvements over the baseline LVViT model are similar, with SaiT-LV-S slightly higher. Note the throughputs for EViT are from the original paper measured on one A100 machine. We could not find the pretrained EViT-LV-S in time for direct comparison at the same machine. 

Same as SaiT-S, one single SaiT-LV-S model achieves competitive accuracy and throughput as two EViT-LV-S models optimized individually at 0.5 and 0.7 keep rates.  

\begin{table}[ht]
  \caption{Performance comparison of EViT-LV-S and SaiT-LV-S. Note EViT-LV-S* are two separated models trained with keep rates at 0.7 and 0.5, whereas SaiT-LV-S/V0.5 is one unified model, trained once with value-based pruning at 0.5 token density and inferred at various token densities. Throughputs of SaiT-LV-S are measured on one 48GB RTX Quadro-8000 GPU with a batch size of 64, following the same procedure in \citep{swin}. $^1$ Throughputs of EViT-LV-S* are from \citep{evit} measured on one A100 GPU.  }
  \label{evit-lv-s vs. sait-lv-s}
  \centering
  \resizebox{\textwidth}{!}{
  \begin{tabular}{cccc|cccc}
    \\
    \multicolumn{1}{c}{\bf Keep rate} 
    &\multicolumn{1}{c}{\bf GFLOPs}
    &\multicolumn{1}{c}{\bf Top-1 Acc.} 
    &\multicolumn{1}{c|}{\bf Throughput} 
    
    &\multicolumn{1}{c}{\bf Token density} 
    &\multicolumn{1}{c}{\bf GFLOPs}
    &\multicolumn{1}{c}{\bf Top-1 Acc.} 
    &\multicolumn{1}{c}{\bf Throughput} 
    
    \\
    \hline
    LVViT &6.6 &83.3 &2112$^1$ &LVViT &6.6 &83.3 &614 \\
    \hline
    \multicolumn{4}{c|}{\bf EViT-LV-S*} & \multicolumn{4}{c}{\bf SaiT-LV-S/V0.5} \\
    \hline
    
    0.7   &4.7 (-29\%)   & 83.0 (-0.3) &2954 (+40\%)$^1$  
    &0.5  &4.3 (-35\%)   & \textbf{83.1 (-0.2)} & \textbf{942 (+53\%)}\\
    
    0.5   &3.9 (-41\%)   & 82.5 (-0.8) &3603 (+71\%)$^1$ 
    &0.4  &3.9 (-41\%)   & \textbf{82.7 (-0.6)} & \textbf{1056 (+72\%)} \\
    
  \end{tabular}
  }
\end{table}

\subsection{Token sparsification sensitivity of DeiT-S}

\begin{figure}[ht]
  \centering
  \includegraphics[scale=0.5]{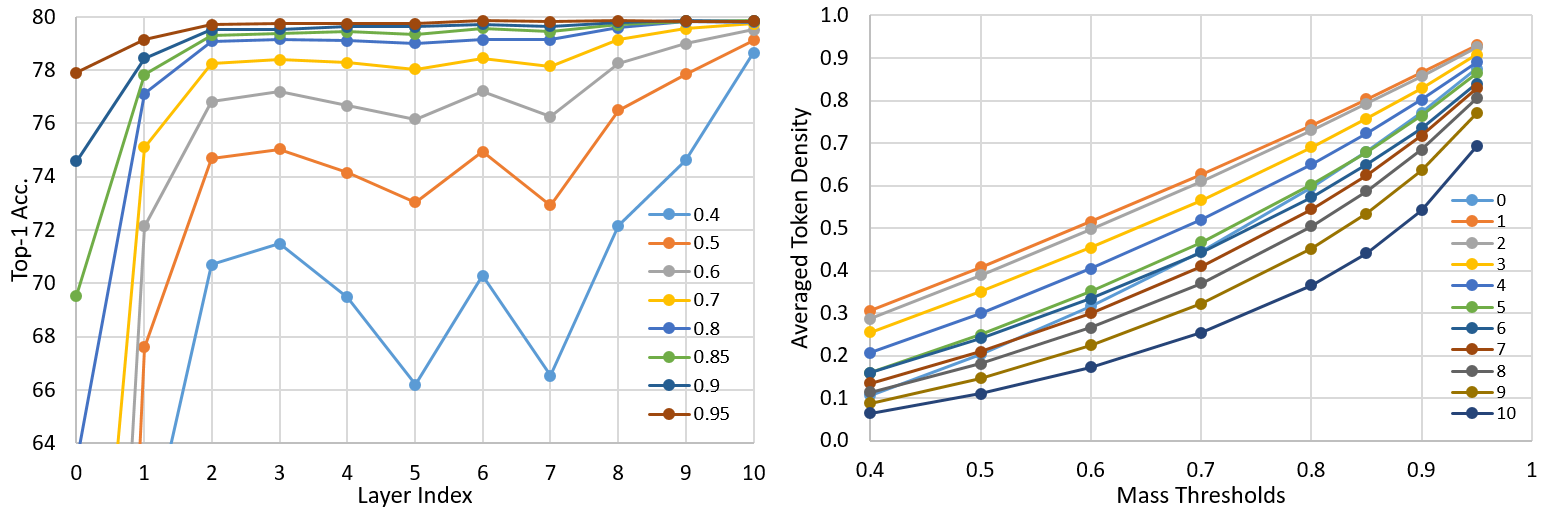}
  \caption{a) Token sparsification sensitivity of DeiT-S: top-1 accuracy when pruned at Layer 0 to Layer 10 with various mass thresholds 0.4 - 0.95. b) Averaged token density with various mass thresholds when pruned at Layer 0 to Layer 10.}
  \label{sensitivity}
\end{figure}

Finding the ideal pruning location needs to balance the following two effects. The earlier the pruning starts, the more computation savings are achieved. However, the capacity to learn $TIS$ is constrained if there are too few transformer modules before the pruning location. To address this problem, we analyze the token sparsification sensitivity of DeiT-S as shown in Figure \ref{sensitivity}. 

For the mass-based sparsification scheme, starting at the first two layers leads to significant accuracy drop (Figure~\ref{sensitivity} a). Layer 3 presents a local optimum, outperforming Layer 2 and Layer 4-5 at various mass thresholds. We hypothesize this results from the $TIS$ distribution along the depth direction as shown in Figure \ref{sensitivity} b). With the same mass threshold, the token density decreases monotonically from Layer 1 to Layer 10. Therefore, we choose Layer 3 as the pruning location in SaiT-S, balancing the accuracy drop and computation savings. Similarly, Layer 3 and Layer 5 are chosen as the pruning locations for SaiT-LV-S and SaiT-LV-M, respectively. Token sparsification sensitivity analysis of LVViT is included in the supplemental materials. Empirically, we find that the optimal pruning layer locates at around Layer $L/4$.

\subsection{Token sparsification sensitivity of LVViT-S}

\textbf{Mass-based sparsification strategy}

\begin{figure}[ht]
  \centering
  \includegraphics[scale=0.5]{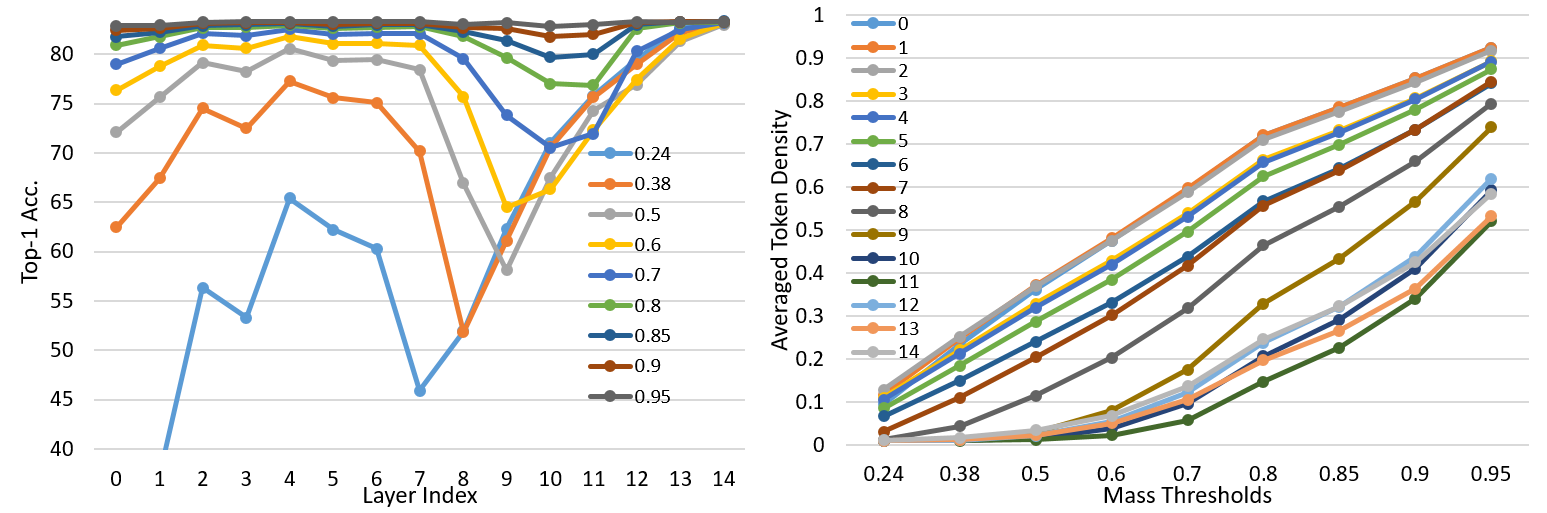}
  \caption{a) Mass-based token sparsification sensitivity of LVViT-S: top-1 accuracy when pruned at Layer 0 to Layer 14 with various mass thresholds ranging from 0.24 to 0.95. b) Averaged token densities with various mass thresholds when pruned at Layer 0 to Layer 14.}
  \label{sensitivity_lvvit}
\end{figure}

As with DeiT-S, the first four layers of LVViT are most sensitive to pruning with significant accuracy drop. As shown in Figure~\ref{sensitivity_lvvit} a), Layer 4 is a local optimum for various mass thresholds. Therefore, we choose Layer 4 as the token pruning location for the mass-based sparsification scheme. 

However, unlike DeiT-S, under the same mass threshold, the token density of LVViT-S no longer decreases monotonically from Layer 1 to Layer 14 (Figure~\ref{sensitivity_lvvit} b). Instead, Layer 10 and Layer 11 show lower average token density than Layer 13 and Layer 14 under the same mass threshold. 
We hypothesize this results from the auxiliary dense patch token labels used by LVViT. Besides learning from the classification label with the classification token, LVViT also leverages the feature representation from all the patch tokens and learns from the patch labels. Therefore, the probability mass at the last few layers is re-distributed across all the patch tokens, as opposed to being more concentrated in the $CLS$ token as in DeiT. 

\begin{figure}[ht]
  \centering
  \includegraphics[scale=0.52]{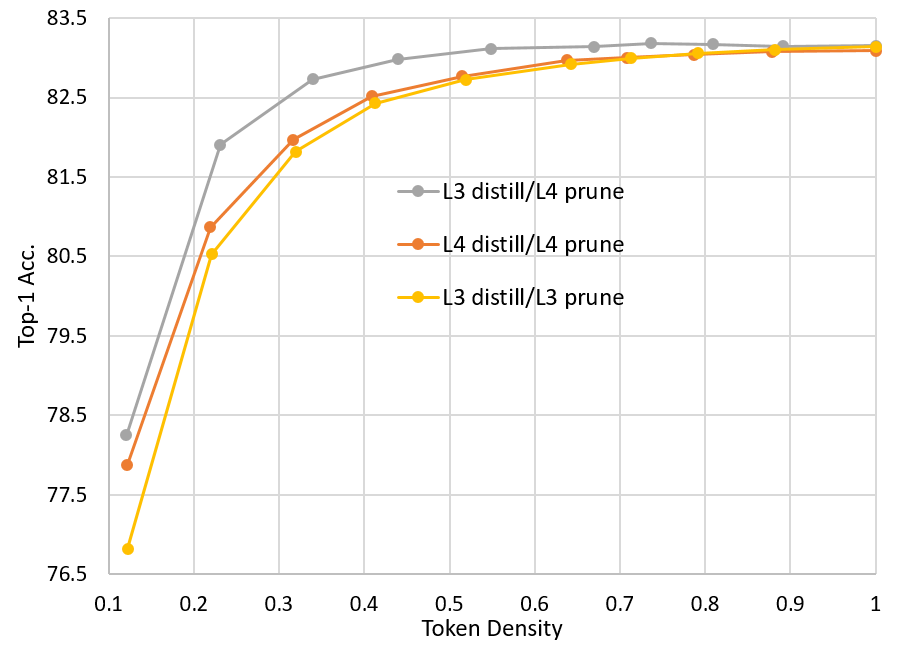}
  \caption{Top-1 accuracy of SaiT-LV-S/M0.7 trained with different knowledge distillation locations and pruning locations. L3 distill/L4 prune denotes the knowledge distillation at Layer 3 and pruning location at Layer 4, while L3 distill/L3 prune represents both knowledge distillation and pruning location at Layer 4, and the same applies to L4 distill/L4 prune.}
  \label{mass_layer}
\end{figure}

Using auxiliary labels for all patch tokens also presents unique challenges for knowledge distillation with LVViT, since the classification token from the last layer only contains partial information regarding token importance. To address this issue, we find that performing distillation at Layer 3 and pruning at Layer 4 yields the best performances as shown in Figure~\ref{mass_layer}. Changing the distillation and pruning location from Layer 3 to Layer 4 only slightly improves the accuracy for averaged token density below 40\%. On the other hand, moving the knowledge distillation location to Layer 3 and keeping the pruning location at Layer 4 significantly improves the accuracy for a wide range of token densities between 0.1 and 0.8. We hypothesize this change injects knowledge from the classification token of the last layer while also allowing more pruning flexibility at Layer 4, which leads to higher accuracy. 

\textbf{Value-based sparsification strategy}

\begin{figure}[ht]
  \centering
  \includegraphics[scale=0.45]{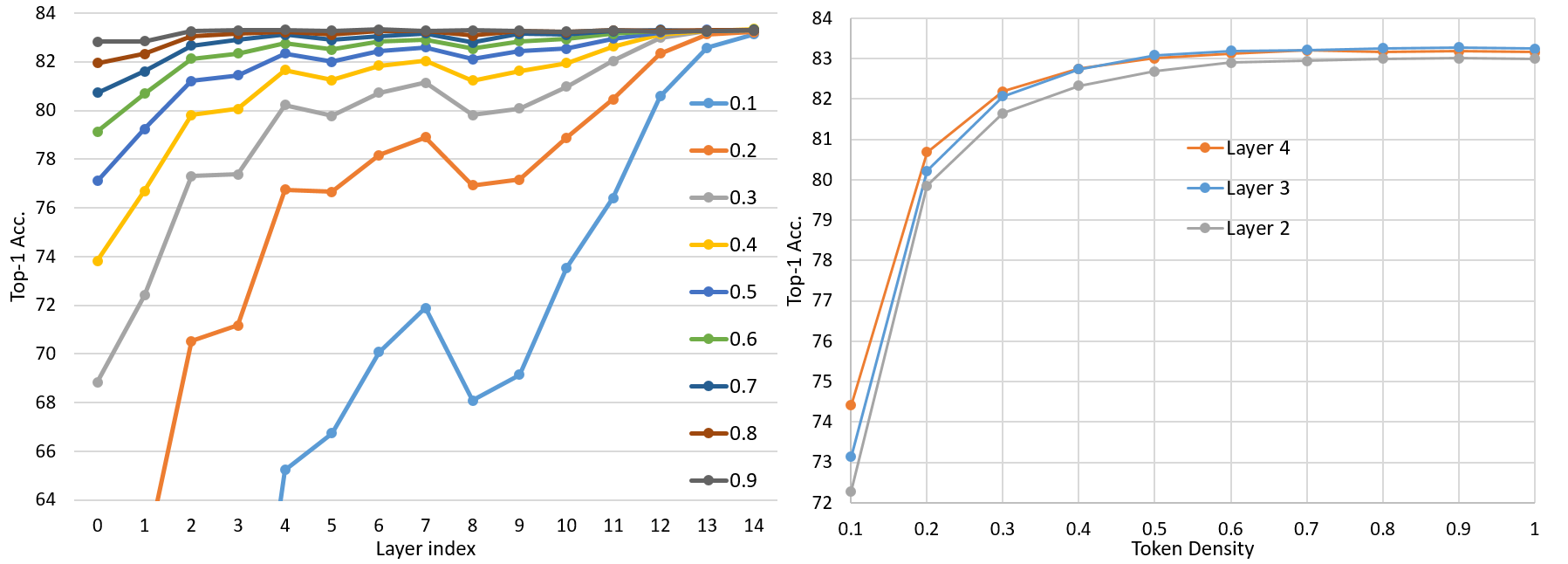}
  \caption{a) Value-based token sparsification sensitivity for LVViT-S: top-1 accuracy when pruned at Layer 0 to Layer 14 with various token densities 0.1 to 1.0 during inference. b) Top-1 accuracy of SaiT-LV-S/V0.5 when training with pruning location at Layer 2, Layer 3, and Layer 4 respectively.}
  \label{value_based_sensitivity}
\end{figure}

Similarly, for value-based sparsification (Figure~\ref{value_based_sensitivity} a), Layer 4 is a local optimum for token densities between 0.2 and 0.7. The differences across different layers get smaller with increasing token densities. Layer 8 and Layer 9 present an accuracy dip, compared to Layer 7 and Layer 10. We hypothesize that different token probability distributions across Layer 8 to Layer 14 are the main cause of the accuracy dip. As a result, we choose not to use knowledge distillation for value-based SaiT-LV models. 


As shown in Figure~\ref{value_based_sensitivity} b), there is little difference in top-1 accuracy of SaiT-LV-S/V0.5 trained with Layer 3 and Layer 4 as the pruning location. However, when moving the pruning location to Layer 2, SaiT-LV-S/V0.5 suffers an average accuracy loss of around 0.3-0.4\% for various token densities.
Therefore, to maximize computation savings while maintaining accuracy, we choose Layer 3 as the default pruning location for SaiT-LV-S/V0.5.

\subsection{Number of training epochs}

\begin{figure}[ht]
  \centering
  \includegraphics[scale=0.56]{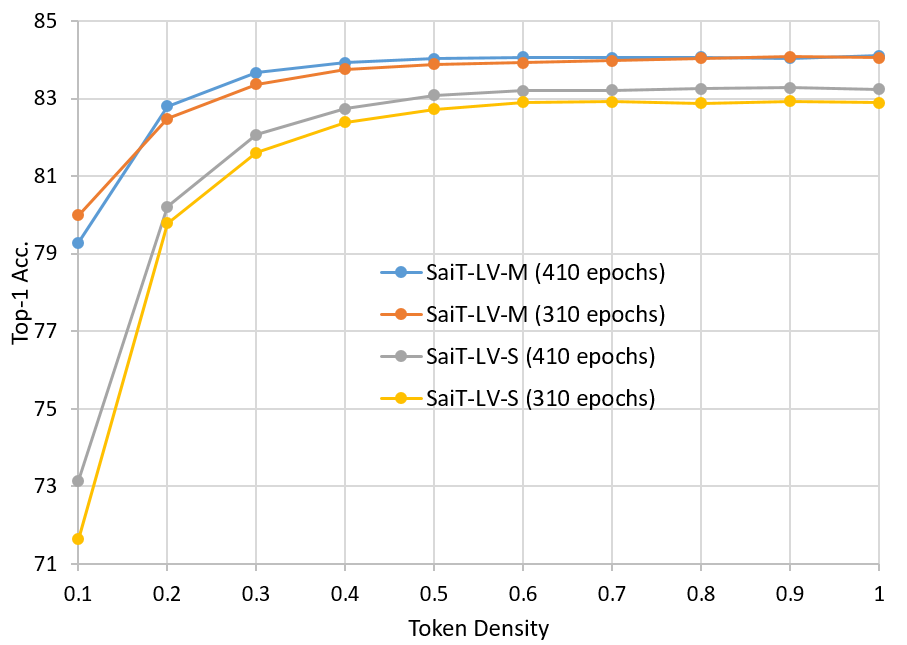}
  \caption{Top-1 accuracy of SaiT-LV-S/V0.5 and SaiT-LV-M/V0.5 trained for 310 epochs and 410 epochs.}
  \label{epochs}
\end{figure}

We compare the performances of SaiT-LV-S/V0.5 and SaiT-LV-M/V0.5 trained for 310 and 410 epochs as shown in Figure~\ref{epochs}. With more parameters and model capacity, the extra 100 epochs of training barely change the accuracy of SaiT-LV-M/V0.5. On the other hand, for SaiT-LV-S/V0.5, 100 more epochs increase the top-1 accuracy by 0.3\% to 0.5\% when inferred at various token densities (0.2 -1.0). As a result, we choose to train the SaiT-LV for 410 epochs to achieve better inference accuracy.

\end{document}